\documentclass{article}

\usepackage{microtype}
\usepackage{graphicx}
\usepackage{subcaption}
\usepackage{booktabs}
\usepackage{hyperref}
\usepackage[table]{xcolor}
\usepackage{amsmath}
\usepackage{amsfonts}
\usepackage{enumitem}

\usepackage[preprint]{icml2026}

\newcommand{\model}{LuckyStar 111B}
\newcommand{\basemodel}{Command A}
\definecolor{tablehighlight}{RGB}{180, 220, 240}

\usepackage{amsmath}
\usepackage{amssymb}
\usepackage{mathtools}
\usepackage{amsthm}

\icmltitlerunning{Think in English, Answer in Korean: Efficient Adaptation of Multilingual Tool-Using Agents}

\begin{document}

\twocolumn[
  \icmltitle{Think in English, Answer in Korean: Efficient Adaptation of Multilingual Tool-Using Agents}

  \icmlsetsymbol{equal}{*}
  \begin{icmlauthorlist}
    \icmlauthor{Utsav Garg}{equal,co}
    \icmlauthor{Sungjin Hong}{equal,co}
    \icmlauthor{Jason Jung}{equal,co}
    \icmlauthor{Justin Lee}{co}
    \icmlauthor{Shaan Desai}{co}
    \icmlauthor{Joon Hee Kim}{co}
    \icmlauthor{Anirudh Shrinivason}{co}
    \icmlauthor{Edmond Wen}{co}
    \icmlauthor{Susie Park}{co}
  \end{icmlauthorlist}
  \icmlaffiliation{co}{Cohere}
  \icmlcorrespondingauthor{SungJin Hong}{sungjinhong@cohere.com}
  \icmlkeywords{agentic AI, multilingual reasoning, tool use, efficient inference, NL2SQL}

  \vskip 0.3in

]

\printAffiliationsAndNotice{\textsuperscript{*}Equal contribution; first authors are listed in alphabetical order.}

\begin{abstract}

We present \textbf{\model{}}\footnote{The name ``LuckyStar'' references the root of the LG brand, ``Lucky Goldstar.''}, a 111B-parameter hybrid reasoning model developed through a collaboration between Cohere and LG CNS for Korean-English enterprise agents under practical memory and serving constraints. The model trains from Cohere's fully post-trained \basemodel{}~\cite{cohere2025command} model rather than a new pretraining run, and uses preamble conditioning to switch between concise non-reasoning behavior and longer tool-oriented reasoning. We study four choices for scaling tool-using agents efficiently: multilingual supervised fine-tuning, reinforcement learning with verifiable rewards for multi-step tool-use tasks, language-consistency rewards for Korean user-facing responses, and 4-bit quantization for single-GPU serving. The adapted model improves mathematical reasoning, function calling, and agentic natural-language-to-SQL (NL2SQL) performance while preserving general Korean and English instruction-following quality. These results provide a practical recipe and failure-mode analysis for adapting post-trained multilingual models to verifiable agentic workflows under memory-constrained deployment.
\end{abstract}

\section{Introduction}

Enterprise assistants must often reason over private data, call tools, and answer under tight memory and serving budgets. Multilingual use makes this harder. A Korean enterprise assistant, for example, may need to parse a Korean request, reason over English-heavy schemas or documentation, execute SQL or retrieval tools, and return a Korean answer whose numbers are grounded in the retrieved or computed result.

We study this setting with \model{}, a 111B-parameter bilingual model jointly built by Cohere and LG CNS for Korean and English enterprise agents. Cohere has previously adapted enterprise models for specific languages and cultures, as in Command R7B Arabic~\cite{alnumay2025commandr7barabicsmall}; here we focus on adding hybrid reasoning and tool-use behavior in a bilingual Korean-English deployment setting. Instead of training a new model from scratch, we adapt a fully post-trained \basemodel{}~\cite{cohere2025command} model. This preserves general instruction-following and multilingual ability while reducing the cost and risk of adding specialized reasoning behavior. The work is motivated by a deployment setting in which agent quality, verifiable tool use, and memory footprint must be considered together.

The main design choice is \emph{hybrid reasoning by preamble conditioning}. The same model weights support two operating modes. A reasoning preamble elicits longer step-by-step behavior for tool use, mathematics, and NL2SQL. A non-reasoning preamble elicits concise responses for ordinary user interactions. This avoids separate models for different enterprise workloads and lets downstream systems choose reasoning depth at inference time.

We make three contributions:
\begin{itemize}[leftmargin=*, itemsep=3pt, topsep=3pt, parsep=3pt, partopsep=1pt]

    \item We describe a three-stage adaptation pipeline for adding reasoning and tool-use behavior to a post-trained multilingual model without full retraining.
    \item We analyze multilingual reasoning failures, especially the tendency for Korean prompts to drift into English final answers, and introduce a reward penalty that improves language consistency during RLVR.
    \item We report deployment-oriented results showing that 4-bit quantization preserves benchmark quality in our evaluations while enabling single-H100 serving for a 111B-parameter model.
\end{itemize}

\section{Method}

\begin{figure*}[t]
  \centering
  \includegraphics[width=\textwidth]{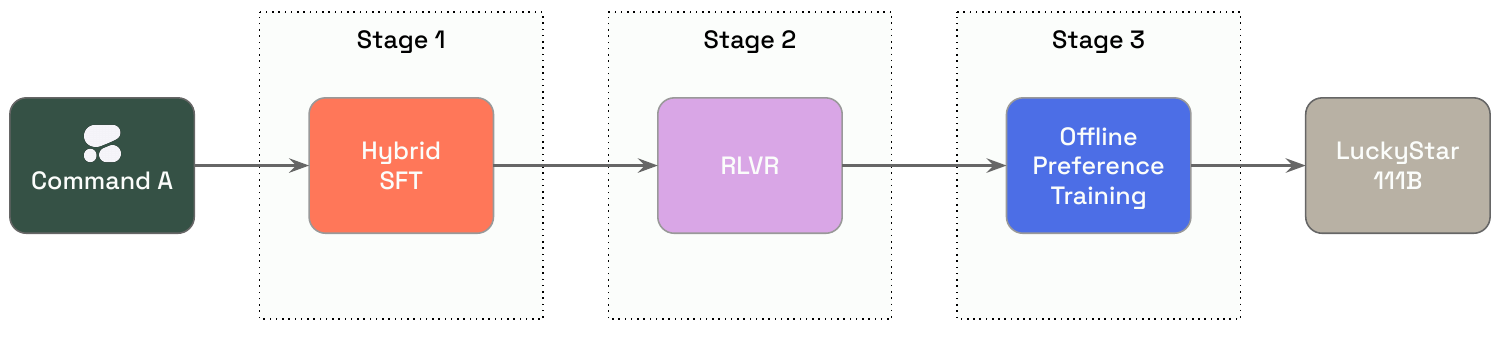}
  \caption{Training pipeline for \model{}. A post-trained multilingual base model is adapted through hybrid SFT, RLVR for verifiable reasoning and tool use, and preference alignment for concise user-facing behavior.}
  \label{fig:overview}
\end{figure*}

\subsection{Hybrid Adaptation Pipeline}

\model{} is initialized from \basemodel{}~\cite{cohere2025command}, a fully post-trained 111B-parameter enterprise model. We add reasoning ability through three stages: hybrid supervised fine-tuning (SFT), reinforcement learning with verifiable rewards (RLVR), and offline preference alignment using Direct Preference Optimization (DPO)~\cite{ouyang2022traininglanguagemodelsfollow,deepseekai2025,rafailov2023direct}.

Hybrid SFT teaches the two modes. Roughly 80\% of the SFT mixture contains reasoning examples, while 20\% preserves non-reasoning instruction-following behavior. The mix is biased toward reasoning because the base model already has strong multilingual instruction-following behavior. Reasoning examples use a preamble that encourages intermediate reasoning before the answer. Non-reasoning examples use a preamble that encourages concise final responses. At inference time, systems can select the desired behavior without changing weights or architecture.

The reasoning mixture includes math, code, and tool-use data. For math and code, we combine internal examples with public reasoning datasets~\cite{guha2025openthoughtsdatarecipesreasoning,deepmath}. Some sourced problem sets contain prompts and final ground-truth answers but no solution traces. For these cases, we generate candidate step-by-step completions with strong internal and open models~\cite{qwen3,deepseekai2025}. We then extract the predicted final answer and retain only completions that exactly match the known ground truth. When answer extraction is unreliable, an LLM judge compares the predicted answer with the ground truth and checks whether the output follows the expected format. For a single prompt, multiple correct traces are often retained; this teaches the model that several reasoning paths can lead to the same answer and improves variation in reasoning style.

\subsection{Multilingual Reasoning Strategy}

\begin{table}[b]
\centering
\resizebox{\columnwidth}{!}{
\begin{tabular}{lcc}
\toprule
\textbf{Task} & \textbf{English Reasoning} & \textbf{Korean Reasoning} \\
\midrule
AIME 2024 & 69.3 & 50.9 \\
MATH 500 & 93.6 & 85.6 \\
\bottomrule
\end{tabular}
}
\caption{Pass@1 accuracy for Korean prompts when reasoning traces are generated in English or Korean during early SFT experiments.}
\label{tab:lang_reasoning_comparison}
\end{table}

Early experiments showed a consistent gap between English and Korean reasoning traces for the same Korean prompts. As shown in Table~\ref{tab:lang_reasoning_comparison}, English reasoning outperformed Korean reasoning even when the prompt was Korean. We attribute this to the English-heavy reasoning distribution and to tokenizer and reasoning-signal differences that favor English traces.

We first machine-translated 30\% of English reasoning traces into Korean using an internal \basemodel{} configuration (command-a-03-2025, temperature 0.3, $p=0.95$), with translations returned in JSON format for safe parsing. Thirty Korean annotators reviewed the translated samples to check naturalness and reasoning fidelity. This produced little improvement, likely because translation altered the naturalness of the reasoning traces and degraded the learning signal. We therefore used a mixed-language strategy: reason in English, but produce the final answer in the user's language. To teach this behavior, we sampled 20{,}000 Korean prompts with intermediate difficulty, generated English reasoning traces paired with Korean final answers through rejection sampling, and added verified examples to SFT.

\subsection{Agentic NL2SQL Data}

To support agentic reasoning for natural-language-to-SQL tasks, we curated NL2SQL data from Spider, BIRD, and SynSQL~\cite{yu2018spider,li2023can,li2025omnisql}. These datasets provide natural-language questions, gold SQL queries, and corresponding SQLite databases. Many raw examples were unsuitable for automated verification: some gold SQL queries failed, some returned empty outputs, and others produced very long results. For RLVR and rejection sampling, we needed examples that could be checked automatically, so we retained only gold queries that executed successfully and returned outputs between one and 500 tokens. This yielded 100{,}000 verifiable prompt-query pairs.

Enterprise tools are a natural setting for verifiable rewards because many actions produce structured or executable outputs. SQL execution, retrieval-grounded answer checks, and formatting constraints can provide sharper training signals than preference-only supervision for workflows where correctness is operationally important.

\begin{figure}[t]
  \centering
  \includegraphics[width=\columnwidth]{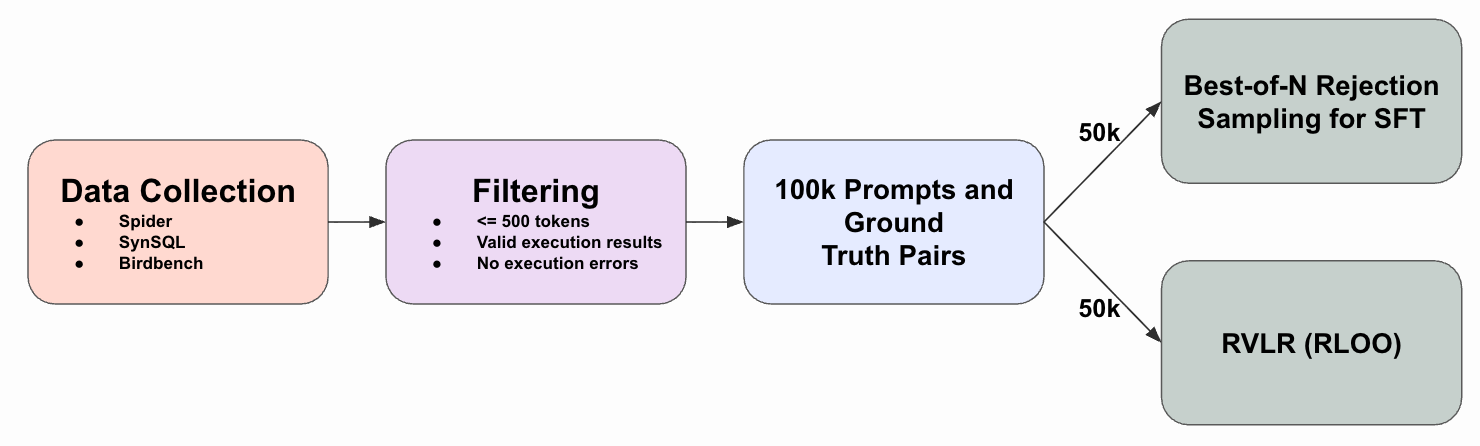}
  \caption{NL2SQL data preparation. Executable gold queries and verified model-generated alternatives provide rewardable tool-use examples.}
  \label{fig:nl2sql_overview}
\end{figure}

The base policy achieved less than 1\% accuracy on this curated NL2SQL set, creating a cold-start problem with insufficient reward signal for reinforcement learning. We first bootstrapped the policy with SFT examples generated by best-of-$N$ rejection sampling~\cite{verdun2025softbestofnsamplingmodel}. A subset of prompts was held out from the RLVR stage to preserve starting entropy. A high-performing internal Cohere model and open models received a read-only SQL execution tool and generated eight candidate solutions per prompt. Candidate queries were executed and verified with a two-stage pipeline combining heuristic checks and an LLM judge for ambiguous cases. We retained all verified-correct traces, including multiple diverse solutions for the same prompt, then integrated the resulting agentic NL2SQL examples into the hybrid SFT mixture.

\subsection{RLVR and Language Consistency}

\begin{table}[b]
\centering
\begin{tabular}{lc}
\toprule
\textbf{Training Step} & \textbf{Language Mismatch Rate} \\
\midrule
Step 0 (SFT) & 12.2\% \\
Step 100 & 3.3\% \\
Step 200 & 0.8\% \\
\bottomrule
\end{tabular}
\caption{Final answer language mismatch rate for Korean prompts during RLVR.}
\label{tab:language_switching}
\end{table}

The RLVR stage optimizes reasoning and tool use with binary rewards. We use REINFORCE Leave-One-Out (RLOO)~\cite{ahmadian2024back} and custom reward functions for mathematics and agentic NL2SQL. The training data contains over 50{,}000 agentic NL2SQL prompts and 4{,}000 mathematics prompts. For each task, the final answer is evaluated against a ground truth using a rule-based verifier when possible. When deterministic rules are insufficient, an LLM judge based on \basemodel{} assigns a binary reward. To optimize training, we select prompts whose pre-RLVR pass rate is between 12.5\% and 87.5\%, avoiding examples that are either solved too easily or provide no reward signal.

For Korean prompts, early RLVR runs often optimized toward English final answers. We therefore use a language-consistency penalty:

\setlength{\abovedisplayskip}{3pt}
\setlength{\belowdisplayskip}{6pt}
\setlength{\abovedisplayshortskip}{3pt}
\setlength{\belowdisplayshortskip}{6pt}
\begin{equation}
\begin{aligned}
r &= r_{\mathrm{correctness}} - r_{\mathrm{penalty}},\\
r_{\mathrm{penalty}} &=
\begin{cases}
0.5 & \text{if final answer language mismatches,}\\
0 & \text{otherwise.}
\end{cases}
\end{aligned}
\end{equation}

Prompt languages are pre-classified, and ambiguous prompts are excluded. An LLM judge checks whether the final user-facing response matches the prompt language while the intermediate reasoning remains in English. As shown in Table~\ref{tab:language_switching}, this reduces Korean-to-English answer drift during RLVR.

To avoid rewarding premature termination, rollouts exceeding 32k tokens are filtered from training batches rather than assigned zero reward. This choice follows the observation that complex reasoning can require long chains of thought, so a zero reward for length can incorrectly teach the model to stop early. RLVR led to improvements in reasoning and agentic NL2SQL, and we also observed fewer repetition-related failures. However, this improvement came at a cost to conciseness: after RLVR, the model frequently failed to follow instructions requiring brief responses, especially in multiple-choice settings where it would elaborate instead of answering directly.

\subsection{Offline Preference Alignment}

To correct the verbosity introduced during RLVR, we apply low-learning-rate DPO over preference pairs~\cite{rafailov2023direct}. The preference set combines data from the base model's alignment pipeline with newly curated Korean-specific preference data. We also generate targeted preference pairs for conciseness: direct answers are labeled as chosen, while verbose post-RLVR outputs are labeled as rejected. The low learning rate is used to refine chatbot behavior without compromising reasoning and tool-use capabilities acquired during RLVR.

\section{Results}

\begin{table*}[t]
\centering
\resizebox{\textwidth}{!}{
\begin{tabular}{p{112pt}cccc|ccc}
\toprule
& \multicolumn{4}{c|}{\textbf{Mathematical Reasoning}} & \multicolumn{3}{c}{\textbf{Tool Use}} \\
\cmidrule(lr){2-5} \cmidrule(lr){6-8}
& \multicolumn{2}{c}{\textbf{Korean}} & \multicolumn{2}{c|}{\textbf{English}} & \textbf{Enterprise NL2SQL} & \textbf{BFCL v3} & \textbf{LG Agentic Eval} \\
& AIME 2024 & MATH 500 & AIME 2024 & MATH 500 & & & \\
\midrule
\rowcolor{tablehighlight}
\textbf{\model{}} & 69.3 & 93.6 & 73.7 & 94.0 & 38.0 & 70.6 & 4.85 \\
\textbf{\model{} 4-bit}\footnotemark & 66.0 & 95.6 & 72.3 & 96.0 & -- & -- & -- \\

\midrule
\textbf{\basemodel{}} & 10.0 & 76.2 & 13.3 & 79.6 & 7.3 & 64.4 & 2.67 \\
\textbf{GPT-4o (11/20)} & 16.7 & 76.4 & 13.3 & 78.6 & 37.3 & 71.7 & 2.56 \\
\textbf{Claude 3.7 Sonnet} & 46.7 & 92.8 & 50.0 & 92.5 & 38.0 & 58.6 & 5.04 \\
\textbf{Qwen3 235B A22B} & 80.0 & 92.4 & 80.0 & 91.2 & 22.0 & 67.9 & 3.29 \\
\bottomrule
\end{tabular}
}
\caption{Reasoning and tool-use results. Mathematical tasks report Pass@1 accuracy.}
\label{tab:reasoning_agent}
\end{table*}
\footnotetext{Tool-use results are not reported for the 4-bit model because the evaluation runs were not completed before the final evaluation cutoff.}

\begin{table*}[t]
\centering
\resizebox{\textwidth}{!}{
\begin{tabular}{p{112pt} *{4}{c} *{4}{c}}
\toprule
& \multicolumn{4}{c}{\textbf{Korean}} & \multicolumn{4}{c}{\textbf{English}} \\
\cmidrule(lr){2-5} \cmidrule(lr){6-9}
& KMMLU & ARC-C & IFEval & MT-Bench & MMLU & ARC-C & IFEval & MT-Bench \\
\midrule
\rowcolor{tablehighlight}
\textbf{\model{}} & 68.6 & 89.2 & 77.8 & 8.09 & 82.7 & 93.8 & 89.4 & 8.45 \\
\textbf{\model{} 4-bit} & 67.9 & 88.6 & 78.3 & 8.06 & 82.1 & 92.8 & 89.3 & 8.50 \\
\midrule
\textbf{\basemodel{}} & 64.9 & 88.8 & 77.2 & 8.18 & 83.9 & 93.4 & 90.9 & 8.34 \\
\textbf{GPT-4o} & 66.3 & 90.5 & 70.2 & 8.73 & 83.7 & 91.8 & 83.0 & 8.53 \\
\textbf{Claude 3.7 Sonnet} & 67.9 & 89.0 & 74.9 & 8.81 & 84.6 & 95.0 & 87.9 & 8.47 \\
\textbf{Qwen3 235B A22B} & 67.0 & 90.3 & 71.3 & 8.87 & 83.4 & 93.2 & 86.4 & 8.73 \\
\bottomrule
\end{tabular}
}
\caption{General Korean and English benchmark performance. KMMLU and MMLU evaluate knowledge; ARC-C evaluates science reasoning; IFEval measures instruction following; MT-Bench is a conversation-quality judge benchmark~\cite{son2024kmmlu,hendrycks2020measuring,clark2018think,zhou2023instruction,zheng2023judging}.}
\label{tab:academic_benchmarks}
\end{table*}

\subsection{Reasoning and Tool Use}

Table~\ref{tab:reasoning_agent} compares \model{} against the base model and strong proprietary and open models, including GPT-4o~\cite{openai2024gpt4ocard}, Claude 3.7 Sonnet~\cite{claude3.7}, and Qwen3 235B A22B~\cite{qwen3}. \model{} improves over \basemodel{} on Korean and English mathematical reasoning, BFCL v3 function calling~\cite{patil2025bfcl}, and internal enterprise tool-use evaluations. For mathematical reasoning, we evaluate AIME 2024~\cite{AIME2024} and MATH 500~\cite{hendrycksmath2021,lightman2023let} in English and Korean-translated form, using a 32k-token maximum context window for all models.

All proprietary-model baselines reflect the evaluation snapshot available at the time these experiments were run. We did not re-run newer model releases after the evaluation freeze.

We evaluate agentic and tool-calling capabilities using public and internal benchmarks. BFCL v3 measures function-calling behavior. Enterprise NL2SQL is an internal Cohere 150-question NL2SQL benchmark spanning multiple business domains. The LG Agentic Evaluation, developed by LG CNS, contains 60 business and finance questions requiring retrieval, SQL, or both; outputs are scored for numerical accuracy, formatting, and intent satisfaction. We include internal enterprise evaluations because public tool-use benchmarks do not fully capture the workflows, schema and formatting requirements, and numerical accuracy constraints that determine whether an agent is useful in deployment. These targeted benchmarks provide strong development signal for fast iteration and for scaling model capabilities toward the final enterprise goal, while public benchmarks provide comparability. 

\model{} improves Enterprise NL2SQL from 7.3 to 38.0 and LG Agentic Evaluation from 2.67 to 4.85. The gains over \basemodel{} are large, but the absolute scores show that realistic enterprise tool use remains difficult.

\subsection{General Multilingual Quality}

Table~\ref{tab:academic_benchmarks} shows that the specialized reasoning pipeline preserves most general instruction-following and multilingual quality. \model{} maintains strong Korean and English performance on knowledge, reasoning, and instruction-following benchmarks. Korean ARC, IFEval, and MT-Bench are translated versions of the English evaluations. On Korean benchmarks, \model{} reaches 68.6 on KMMLU, 89.2 on ARC Challenge, 77.8 on IFEval, and 8.09 on MT-Bench. In English, it reaches 82.7 on MMLU, 93.8 on ARC Challenge, 89.4 on IFEval, and 8.45 on MT-Bench. These results indicate that the reasoning and tool-use gains do not come at the cost of broad Korean and English instruction-following quality.

\subsection{Efficient Deployment}

Enterprise deployments are constrained by inference cost, memory, and available GPU capacity. We quantize \model{} from FP8 to 4-bit weights, reducing memory footprint by approximately 50\% and lowering memory-bandwidth requirements. This enables the 111B-parameter model to run on a single 80GB H100 GPU, lowering the barrier to deployment in enterprise settings. As shown in Tables~\ref{tab:reasoning_agent} and~\ref{tab:academic_benchmarks}, the 4-bit model closely matches the FP8 model across the reported reasoning and academic benchmarks. We view this result as a deployment feasibility check rather than a complete serving study; detailed latency, throughput, and energy measurements remain important future work.

\section{Discussion}

These results suggest that multilingual enterprise agents can benefit from separating the internal reasoning language from the user-facing response language. In our experiments, English reasoning traces provide stronger learning signals for verifiable reasoning, while language-consistency rewards keep Korean prompts from producing English final answers. This design is practical but imperfect: native Korean reasoning remains a target for future work, especially for tasks where cultural or domain-specific knowledge may be better expressed in Korean.

The NL2SQL experiments also show the importance of cold-start data for tool-use RL. In the curated NL2SQL set, the base model produces too few correct traces for RLVR alone. Rejection-sampled SFT examples create enough initial competence for RL to improve policy. This is consistent with the broader observation that reward quality and prompt selection are central to making verifiable RL useful for agentic tasks.

During development, three failure modes were especially important: NL2SQL cold-start failures, language drift from Korean to English during RLVR, and verbosity after reasoning optimization. The final pipeline addresses these with rejection-sampled SFT, a language-consistency reward, and preference alignment. We believe these failure modes are broadly relevant to scalable agentic systems because they arise at the interface between verifiable rewards, multilingual behavior, and deployment-oriented model adaptation.

More broadly, our results support a deployment-scoped view of scaling: practical agent systems do not always require a single model optimized for every capability. For many enterprise settings, it can be more useful to adapt an existing post-trained model toward a focused set of verifiable workflows, while preserving general instruction-following behavior and reducing serving requirements through quantization.

This work has several limitations. Some evaluations are internal because they reflect deployment-specific workflows, schema conventions, and formatting requirements that are not fully captured by public benchmarks; these benchmarks are useful for development but should complement, not replace, public evaluations. Details about the proprietary data and internal evaluation sets are necessarily limited due to legal, privacy, and confidentiality obligations, which also limits reproducibility. Finally, the model still relies on English reasoning for many Korean prompts, leaving native Korean reasoning as an important direction for future work.

\section{Conclusion}

We presented \model{}, a bilingual tool-using agent model adapted from a post-trained 111B-parameter base. Preamble conditioning lets one model support reasoning and non-reasoning modes. Verifiable NL2SQL rewards, language-consistency penalties, and preference alignment improve tool use while preserving Korean and English instruction-following behavior. Finally, 4-bit quantization makes the model deployable on a single H100 with little observed quality loss in the reported evaluations. Future work will focus on expanding native-language reasoning data, improving Korean domain adaptation, measuring serving efficiency more comprehensively, and integrating the model more closely with enterprise knowledge bases and tools.

\section*{Acknowledgments}

This work was a collaboration between many teams at Cohere and LG CNS. We particularly appreciate the members of the \textbf{GenAI Product Lab} and \textbf{Agentic AI Lab} from \textbf{LG CNS AI Lab}. We also acknowledge the following people at Cohere who supported the project: Neeral Beladia, Andrew Chang, Eugene Cho, Q Cho, Elliott Choi, Ali Edalati, Manoj Govindassamy, Vi Iyengar, Edward Kim, Jiyeon Lee, Jeffrey Li, Jonathan Li, Olivia Markham, Adrien Morisot, Vivek Muppalla, Jeremy Pekmez, Max Pfeifer, Sudip Roy, Michael Sachs, Isha Satyakam, and Tom Sherborne.

\section*{LLM/Agent Usage Disclosure}

The research described in this paper used LLMs as part of the model-development pipeline: generating candidate reasoning traces, judging answer equivalence for examples where deterministic verification was insufficient, checking final answer language consistency, and producing candidate SQL traces with a read-only execution tool. LLM was used to help proof-read the paper to improve clarity and flow. The authors are responsible for all technical claims, experiments, and the final text.

\bibliography{references}
\bibliographystyle{icml2026}

\end{document}